\documentclass{article}

     \usepackage[final]{neurips_2020}

\usepackage[utf8]{inputenc} 
\usepackage[T1]{fontenc}    
\usepackage{hyperref}       
\usepackage{url}            
\usepackage{booktabs}       
\usepackage{amsfonts}       
\usepackage{nicefrac}       
\usepackage{microtype}      

\usepackage{times}
\usepackage{latexsym}
\usepackage[pdftex]{graphicx}
\usepackage{bbold}
\usepackage{colortbl}
\usepackage{tabu}
\usepackage{booktabs}
\usepackage{multirow}
\usepackage{subfig}

\usepackage{amsmath}
\DeclareMathOperator*{\argmax}{arg\,max}

\newcommand{\todo}[1]{}
\renewcommand{\todo}[1]{{\color{red} TODO: {#1}}}

\usepackage{microtype}

\title{Decoding and Diversity in Machine Translation}

\author{%
  Nicholas Roberts$^{\spadesuit}$
  \quad 
  Davis Liang$^\diamondsuit$
  \quad
  Graham Neubig$^{\spadesuit}$ 
  \quad
  Zachary C. Lipton$^{\spadesuit}$$^\diamondsuit$ \\
  $^{\spadesuit}$Carnegie Mellon University; $^\diamondsuit$Amazon AWS AI \\
  \texttt{\{ncrobert, gneubig, zlipton\}@cs.cmu.edu}, \texttt{liadavis@amazon.com}
}

\usepackage[compact]{titlesec}
\titlespacing{\subsection}{0pt}{*1}{*0}
\usepackage[subtle, mathdisplays=tight, charwidths=tight, leading=normal]{savetrees}
\def\setstretch#1{\renewcommand{\baselinestretch}{#1}}
\begin{document}

\maketitle

\begin{abstract}
Neural Machine Translation (NMT) systems are typically evaluated 
using automated metrics that assess the agreement 
between generated translations and ground truth candidates. 
To improve systems with respect to these metrics, 
NLP researchers employ a variety of heuristic techniques,
including searching for the conditional mode (vs. sampling)
and incorporating various training heuristics (e.g., label smoothing).
While search strategies significantly improve BLEU score,
they yield deterministic outputs 
that lack the diversity of human translations.
Moreover, search can amplify 
socially problematic biases in the data,
as has been observed in machine translation of gender pronouns.
This makes \emph{human-level BLEU} a misleading benchmark;
modern MT systems cannot approach human-level BLEU
while simultaneously maintaining human-level translation diversity.
In this paper, we characterize \emph{distributional differences}
between generated and real translations,
examining the cost in diversity 
paid for the BLEU scores enjoyed by NMT.
Moreover, our study implicates search as a salient source of
known bias when translating gender pronouns.  
\end{abstract}

\section{Introduction}
\label{sec:intro}
Neural Machine Translation (NMT)
typically proceeds in the following two-stage pipeline: 
\textbf{(i)} train a conditional language model (using neural networks) 
by optimizing a probabilistic objective (the \emph{modeling} stage);
then \textbf{(ii)} produce predictions by \emph{searching}
for the mode (the hoped-for ``best translation'') 
of the conditional distribution
(decoding either greedily or via beam search).
We confirm that search 
is remarkably effective at maximizing BLEU.
In fact, an NMT model trained for only $1/3$ of an epoch 
and decoded via search can match the BLEU score 
of a fully trained model decoded via sampling.
Moreover, the fully trained model gains an additional 14 BLEU points 
when we decode deterministically via search instead. 
%
%
%

%
%
%
%

However, due to search (either beam search or greedy decoding), 
NMT models are dialed to an extreme operating point of
exhibiting zero variability (conditional on input)
whereas multiple human translations
exhibit considerable variability. 
Yet NMT systems are often rated 
against ``human-level'' performance 
(calculated via BLEU on sentences with 
multiple available translations \citep{ott2018analyzing}), which makes
this comparison misleading. 
Diversity in NMT is valuable for numerous reasons.
For example, homogeneity can make language generation outputs monotonous and less engaging to users. In addition, another pernicious problem that we stress in this paper is that individuals who interact with language primarily through NMT might develop a warped exposure to that language. As one specific example of this, we demonstrate that even when translating between two gendered languages, search will \emph{disproportionately} choose the more frequent gender, conditioned on the input.
We stress that these issues are inherent 
to using deterministic search methods, such as beam search and greedy decoding,
to recover high probability translations 
for the sake of optimizing BLEU score. 
In turn, the singular focus on improving BLEU 
leaves no incentive to address issues of diversity. 

In this paper, we expose search as a cause 
of the lack of diversity in NMT outputs, 
as it relates to the translation of gender pronouns 
and a battery of other diversity metrics that we introduce. 
Specifically, we propose a panel 
of diversity diagnostics for NMT systems,
measuring the distributional similarity
(vs ground truth translations)
of $n$-grams, sentence length, punctuation, and copy rates.
We also examine domain confusion scores,
using both linear discriminators 
with term frequency–inverse document frequency (TF-IDF) features
and BERT-based 
discriminators~\citep{devlin-etal-2019-bert} 
to distinguish between generated and real translations. 
Our study centers on the WMT 2017 German-English (De-En), 
English-German (En-De), and WMT 2014 French-English (Fr-En) datasets. 
We note that the focus on metrics like BLEU
can cause researchers to disregard the consequences of 
ad hoc decisions that may improve BLEU 
while undermining any straightforward 
probabilistic interpretations of the learning objective.
Minimizing cross entropy on the original targets
corresponds to maximizing the likelihood of the data,
but what can we say about the parameters that
minimize cross entropy against label-smoothed targets?
Examining the effect of label smoothing on the diversity of NMT outputs,
we find that when sampling, this results in poor performance
in both BLEU score and our diversity diagnostics.
When decoding via beam search
the effect of label smoothing is minimal,
which can lead to the negative consequences 
going unnoticed by the practitioner. 



\begin{figure*}[t!]
  \centering
  \includegraphics[width=0.85\textwidth]{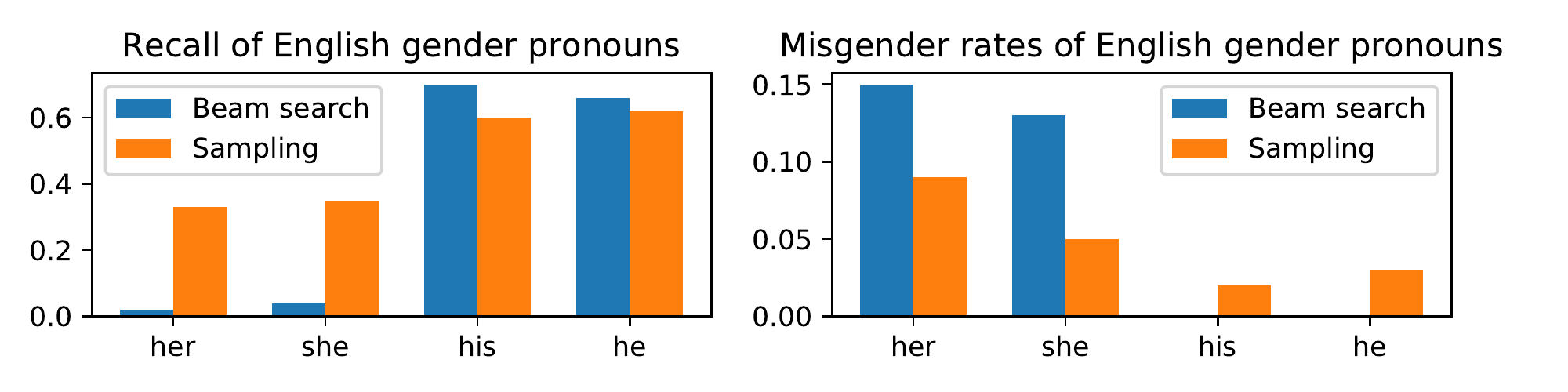}
  \vspace{-7px}
  \caption{Recall and misgender rates of English gender pronouns are made worse by beam search. 
  }
  \vspace{-10px}
  \label{fig:gender}
\end{figure*}

\section{Related Work}
\label{sec:related}
%
%
BLEU score was designed to correlate 
with human judgements of translation quality \citep{papineni2002bleu}, 
although several studies have questioned this correspondence 
\citep{callison-burch-etal-2006-evaluating, ma-etal-2019-results}.
\citet{ott2018analyzing} explores the lack of diversity 
in NMT outputs and relate this to inherent uncertainty in the task.
We note in passing that their study does not document the use of label smoothing,
yet their observation that the ``model distribution is too spread in hypothesis space'' 
is an obvious consequence of label smoothing. 
We consider a broader set of diversity metrics and decoding strategies 
while using the same models and languages to facilitate conversation between these works. 
Furthermore, we explicitly consider the impact of label smoothing. 
Other studies propose decoding strategies to increase diversity,
but lack comparisons to the ground truth distribution
\citep{gimpel-etal-2013-systematic, vijayakumar2016diverse, li2016mutual, cho2016noisy}. 
It has been observed that NMT can often produce misgendered outputs, 
and solutions focusing on the modeling stage have been proposed, 
however, our study implicates search as a source of gender pronoun bias 
\citep{vanmassenhove-etal-2018-getting, saunders-byrne-2020-reducing}.


\citet{tevet2020evaluating} propose a framework intended for tasks outside of NMT for evaluating diversity metrics relative to a ``diversity parameter'' used in decoding, expecting correlation between the metrics and the parameter. 
They consider as metrics the number of distinct $n$-grams from the model output, as done by \citet{li-etal-2016-diversity},
and BERT-based sentence similarity scores. In contrast, we additionally compare against the ground truth distribution of $n$-grams and use BERT as a discriminative model. 

\citet{muller2019} examine the role of label smoothing  
in NMT, claiming that improvements arise
due to improved calibration of the model. 
Notably, \citet{vaswani2017attention} employs label smoothing 
to improve beam search outputs at the expense of perplexity. 
Label smoothing may reduce 
overconfidence of predictions, which can be beneficial in light of known
miscalibration issues in NMT models \citep{kumar2019calibration}.
However, we note here that label smoothing
\emph{is not a valid calibration procedure}
and unsurprisingly, label-smoothed models remain miscalibrated \citep{wang2020inference}. Most importantly, we find label-smoothing negatively impacts various human-level desiderata when sampling. 



\section{Experiments}
\label{sec:experiments}
\noindent \textbf{Implementation details \quad} 
We use the same convolutional architecture 
as \citet{ott2018analyzing} in our experiments. 
For experiments using label smoothing, we set it to $0.1$.
We perform our analysis on the WMT'17 En-De dataset and present results 
in both directions. 
We repeat our analysis on the much larger WMT'14 En-Fr dataset, 
where we fix the task to be translation from French to English 
(the results of which can be found in Appendix~\ref{app:diagnostic}). Additional implementation details can be found in Appendix~\ref{app:implementation}. 
The sampling method used in our experiments involves randomly sampling from the softmax with some temperature at each time step. For beam search, we do not include any additional penalties. For all sampling and search procedures, the output up to the current time step is passed as input to the decoder. We also note that as the sampling temperature approaches $0$, the sampling procedure deterministically selects the $\argmax$ at each time step, which is equivalent to greedy search, and is also equivalent to beam search with a beam width of $1$. 
\begin{figure*}
  \centering
  \includegraphics[width=\textwidth]{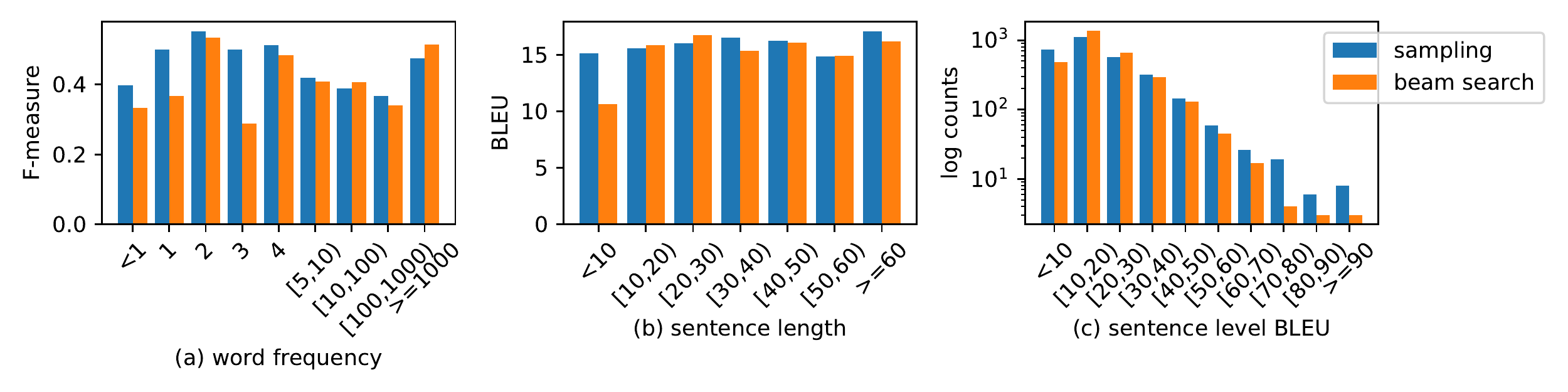}
  \vspace{-15px}
  \caption{
  Subject to the same BLEU score as sampling, beam search 
  (a) has lower F1 scores for rare words, (b) underperforms on short sentences, and (c) has a lower variance in sentence level BLEU scores.
  }
  \vspace{-25px}
  \label{fig:cmpnmt}
\end{figure*}
\vspace{1px}

\noindent \textbf{Beam search is biased towards selecting more frequent gender pronouns.} 
We evaluate the bias of search toward more frequent tokens using a model trained to translate from German to English on the WMT’17 En-De dataset, and draw a direct comparison to sampling (i.e. not performing search at all) by fixing the BLEU score to a particular value for both. Our analysis for this is performed on the test set. 
We reduced the amount of training for the beam search decoded model 
to $1/3$ of an epoch to 
produce a model that achieves \textbf{the same} BLEU score ($16.2$) 
as the fully converged model decoded by sampling. 
We examine distributional differences including correct prediction rates of female pronouns, 
word frequency, sentence length, and the distribution of sentence-level BLEU scores 
between the two systems using compare-mt \citep{DBLP:journals/corr/abs-1903-07926}.
In this setting, we find that beam search underpredicts female gender pronouns. 
We find that the recall for the tokens `she' and `her' were significantly higher when decoding via sampling: $0.35$ and $0.33$, respectively, compared to $0.04$ and $0.02$ from beam search. In contrast, the recall of male pronouns from beam search was higher than that of sampling.
Beam search also replaces `she' and `her' with male pronouns at higher rates than sampling, while never replacing male pronouns with female pronouns (see Figure~\ref{fig:gender}). 
%
We also find that beam search attains a consistently lower F1 for rare words 
compared to sampling (Figure~\ref{fig:cmpnmt}a).
We show in Figure~\ref{fig:cmpnmt}b that beam search achieves lower BLEU scores for shorter sentences, 
and in Figure~\ref{fig:cmpnmt}c, we find that sampling results in higher a variance of BLEU scores than beam search, and in particular, more sentences with BLEU scores of $30$ or higher.

\vspace{5px}
\noindent \textbf{Our diversity diagnostics reveal trade-offs between diversity and BLEU.}
We now focus on fully trained models on translation from English to German and
compare the distributions of translations
produced via beam search and sampling to reference translations.
We present a variety of metrics to capture various aspects of the distributions
and, in the spirit of \citet{tevet2020evaluating}, compare various operating points along a spectrum mediated by dialing the softmax temperature to elucidate the trade-off between diversity and BLEU. 
In particular, we sample using softmax temperatures, $T$, 
ranging from $0$ to $1$ and compare this to beam search with a beam width, $B$, up to $10$. 
We repeat these analyses using models trained with and without label smoothing, 
and note that the side effects of label smoothing are prominent 
when sampling, and minimal when decoding deterministically. 

\begin{figure*}
  \centering
  \includegraphics[width=1.0\textwidth]{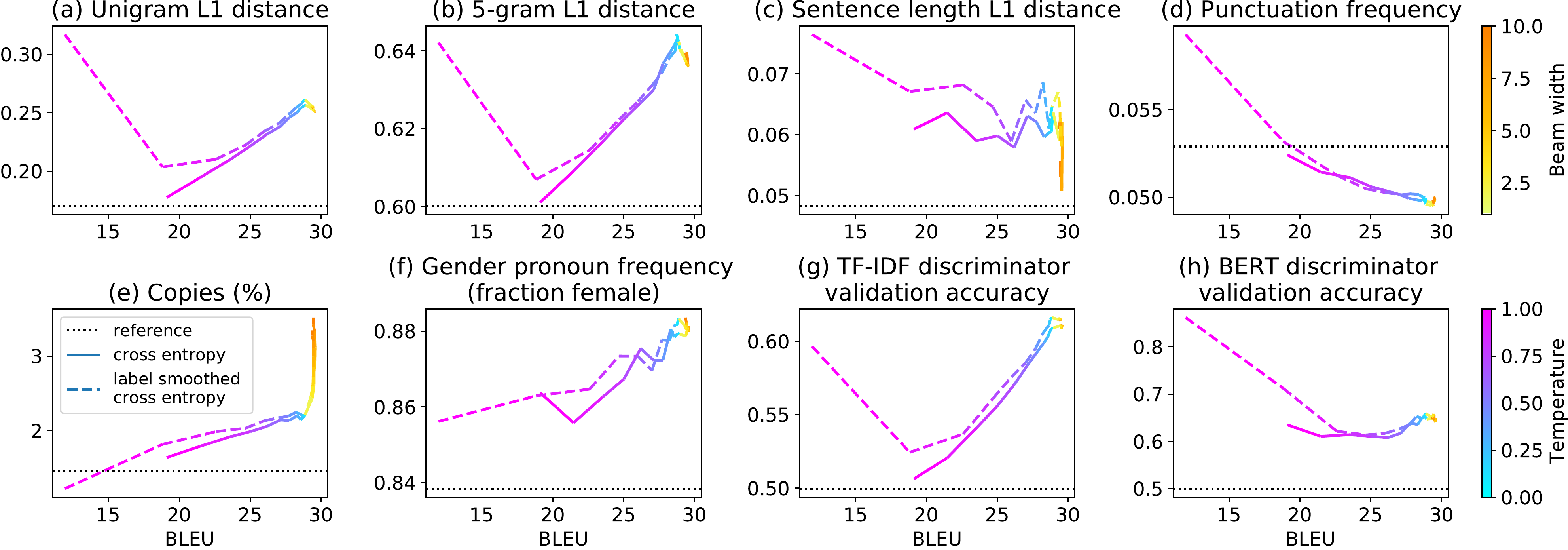}
  \caption{Distributional similarity to ground truth translations across sampling temperatures and beam widths. Improvements in BLEU often come at the cost of distributional dissimilarity. Reference lines are computed between training and validation sets, except for (e) which is the validation copy rate. 
  }
  \vspace{-15px}
  \label{fig:all_sweep}
\end{figure*}

To assess similarity of $n$-gram frequency
and sentence length between generated 
and human translations,
we adopt the L1 distance, 
due to its simplicity and 
robustness to small changes 
in single elements (unlike KL divergence). 
We calculate the L1 distance between 
the normalized histograms
for the model output on the validation set 
for a given decoding strategy
and those of reference translations.
For $n$-grams, we compare frequencies 
of $n$-grams and report results for $n=1,5$. 
When evaluating sentence length, 
all unique sentence lengths are considered.
As a baseline, we compute the L1 distance between validation set partitions. 
We find that a temperature of $1$ achieves L1 distance similar to the baseline 
(Figures~\ref{fig:all_sweep}a and \ref{fig:all_sweep}b). 
For $n$-grams, we observe a trade-off between closeness in L1 distance 
to the ground truth distribution and BLEU score. 
Evaluating the L1 distance between the sentence length distributions,
we find that both sampling and beam search produce outputs with a similar 
distribution of sequence lengths to ground truth candidates.
For $n$-gram L1 distances, sampling results in a closer distributional similarity to the ground truth than beam search. 
In all cases, label smoothing results in a lower distributional similarity to the ground truth (Figure~\ref{fig:all_sweep}a,b,c).

We compare the total frequencies of a subset of the vocabulary in generated text 
to that of validation set references. 
We evaluate the frequencies of punctuation and of male and female gender pronouns (see Appendix~\ref{app:subset}). 
In Figure~\ref{fig:all_sweep}d, we show that punctuation frequency decreases
relative to the reference as temperature decreases, 
and increasing beam width does little to improve this. 
Label smoothing at $T=1.0$ actually increases punctuation frequency 
well above that of the reference. 
In Figure~\ref{fig:all_sweep}f, we evaluate the fraction of pronouns which are female. 
Decreasing temperature from 1 and applying beam search 
increases the fraction of German female gender pronouns. 
The German word `sie' translates as `she,' `they,' or `you' in English,
yielding an effectively higher representation of female gender pronouns than male in the training set. 
Beam search is thus biased toward the more common gender pronoun, `sie.' We see this same effect when translating from German to English and French to English, except in these cases, male pronouns are more represented in the training set. Hence in these cases, the bias is toward male pronouns, as shown in Appendix~\ref{app:diagnostic}. In both cases, beam search outputs are biased toward the more represented gender in the training set. 

We similarly evaluate the rate of copies from the sentences in the source language to the output. 
Like \citet{ott2018analyzing}, we define the copy rate 
as the fraction of sentences with more than 50\% unigram overlap, 
excluding punctuation and numbers. 
We compare these scores with the copy rate between 
the sources and references in the training set. 
Our findings in Figure~\ref{fig:all_sweep}e corroborate 
the findings of \citet{ott2018analyzing} 
regarding the exacerbation of copy rates by beam search. 
We expand on this by showing that sampling at $T=1.0$ results 
in copy rates which nearly match the copy rates measured on the training set. Hence the exacerbation of copy rates is specifically due to beam search. 


Finally, we examine the ability for discriminators to distinguish 
between model outputs on the validation set and validation set reference translations.
%
We first construct a dataset comprising generated translations from the model, labeled `generated,' and reference translations labeled `real,' which are taken from a different partition of the validation set.
We use half of the resulting dataset to train a discriminator 
and we use the remaining half to evaluate the generalization performance of the discriminator. 
We train a logistic regression model using TF-IDF features and, using the same setup, fine-tune a BERT-based discriminator (108M parameters). 
We find that even logistic regression trained on TF-IDF features (Figure~\ref{fig:all_sweep}g) 
can classify samples generated by beam search above 60\% accuracy but cannot distinguish 
between translations generated via sampling with temperature $1$ and reference translations. 
Out of all the methods we evaluated, using label smoothing and
sampling with temperature $1$ produces the most discriminable output. 
A BERT-based discriminator, on the other hand (Figure~\ref{fig:all_sweep}h), 
can distinguish between generated and reference samples better than the linear discriminator. In both cases, outputs generated via sampling are harder to discriminate from ground truth translations compared to outputs generated via search, and outputs from label smoothed models are always easier to classify as `generated.'

\section{Conclusions}
\label{sec:conclusion}
Examining distributional dissimilarity 
between the outputs of NMT systems under sampling, 
beam search, and natural translations,
we find that beam search performs well
by BLEU score,  
but there is a significant cost to be paid in naturalness and diversity, including a higher rate of misgendering of gender pronouns.
Moreover, modifications to the objective 
undertaken to increase BLEU (here, label smoothing),
can have unintended side effects 
that practitioners focused on BLEU might overlook.
In future work, we plan to explore techniques to 
achieve the highest possible BLEU subject to constraints 
on the distributional similarity between generated and natural translations. 

\bibliographystyle{acl_natbib}
\bibliography{anthology,emnlp2020}

\newpage
\appendix

\section{Implementation details}
\label{app:implementation}
The convolutional sequence to sequence model used in our experiments consists of encoder and decoder networks, 
each of which contains several `blocks' of 
convolutional layers \citep{gehring2017convolutional}. 
All models are trained on 8 V100 GPUs using Fairseq-py \citep{ott2019fairseq}, 
with learning rate $0.5$ (for En-De and De-En) and $0.25$ (for Fr-En),
with a fixed learning rate schedule, a clip norm of $0.1$, 
dropout of $0.2$, and $4000$ maximum tokens.
All models were trained to convergence (up to $100$ epochs), 
with the best checkpoint chosen based on validation set performance. 
We note that the sampling method used in our experiments involves randomly sampling from the softmax with a temperature parameter at each time step. For beam search decoding, we do not include any additional penalties on the search. For all sampling and search procedures, the output up to the current time step is passed as input to the decoder. We also note that as the sampling temperature approaches $0$, the sampling procedure deterministically selects the $\argmax$ at each time step, which is equivalent to greedy search, and is also equivalent to beam search with a beam width of $1$.

\section{Token subsets used for punctuation and gender pronoun frequency scores}
\label{app:subset}

\begin{table}[h]
\begin{minipage}{1.0\linewidth}
	\centering
	\footnotesize
\begin{tabu}{@{}lc@{}}
\toprule
\textbf{Category} &  \textbf{Subset}  \\
\midrule
punctuation    & . , ? ! " ' ... !!! ?! !? ; : \\
English female & she, her, hers, herself  \\
English male   & he, him, his, himself  \\
German female  & sie  \\
German male    & er  \\
\bottomrule
\end{tabu}
\end{minipage}
\vspace{10px}
\caption{Subsets used for punctuation and gender pronoun analysis}
\label{table:fequency}
\end{table}

\newpage


\newpage
\begin{figure*}
  \section{Diversity diagnostics applied to De-En and Fr-En}
  \label{app:diagnostic}
  
  \centering
  \vspace{10px}
  \includegraphics[width=1.0\textwidth]{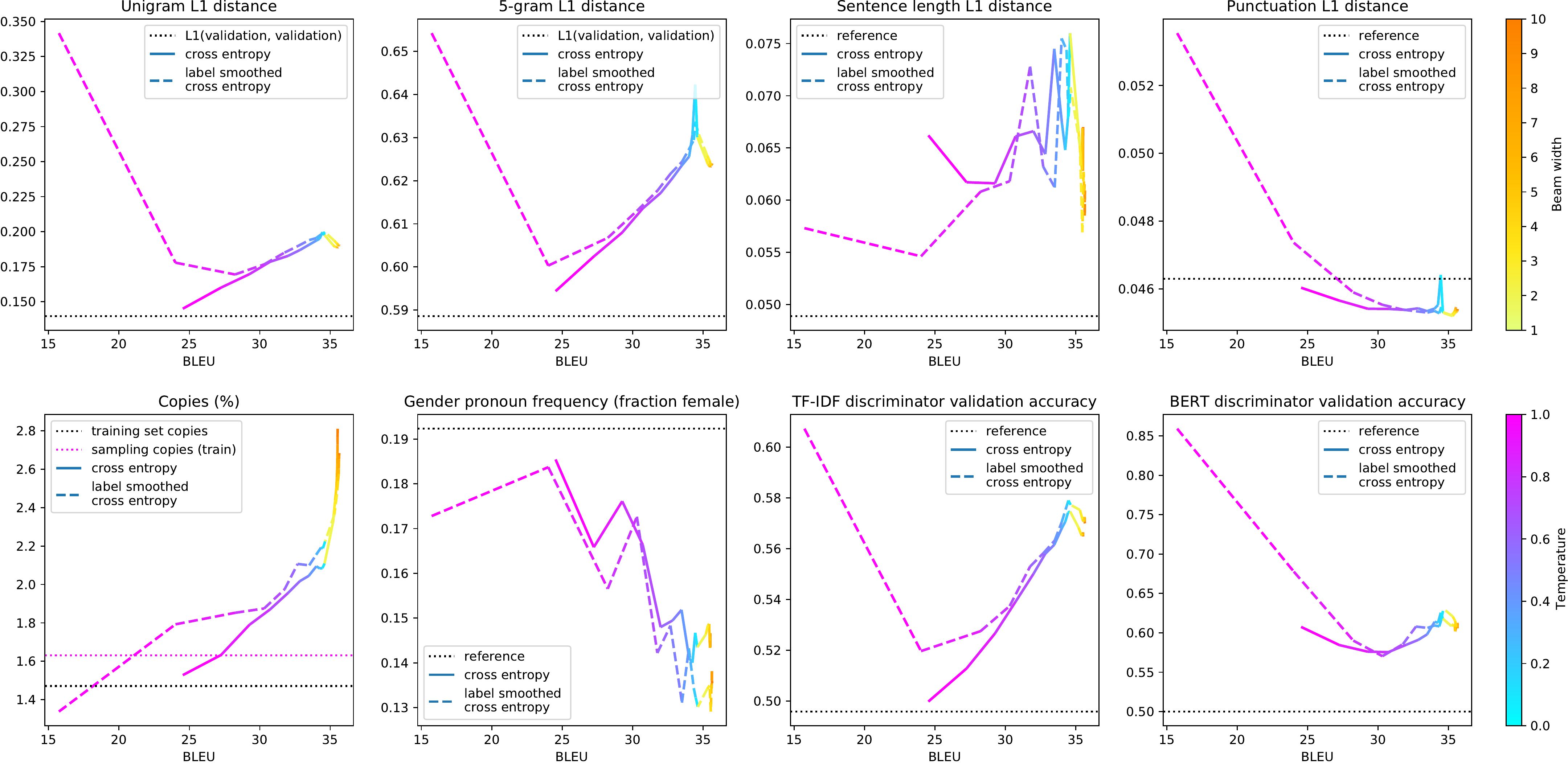}
  \caption{Results from Figure~\ref{fig:all_sweep} reproduced on the WMT 2017 English-German dataset, where the task is translation from German to English. }
  \vspace{10px}
  \label{fig:all_sweep_de_en}
  
  \includegraphics[width=1.0\textwidth]{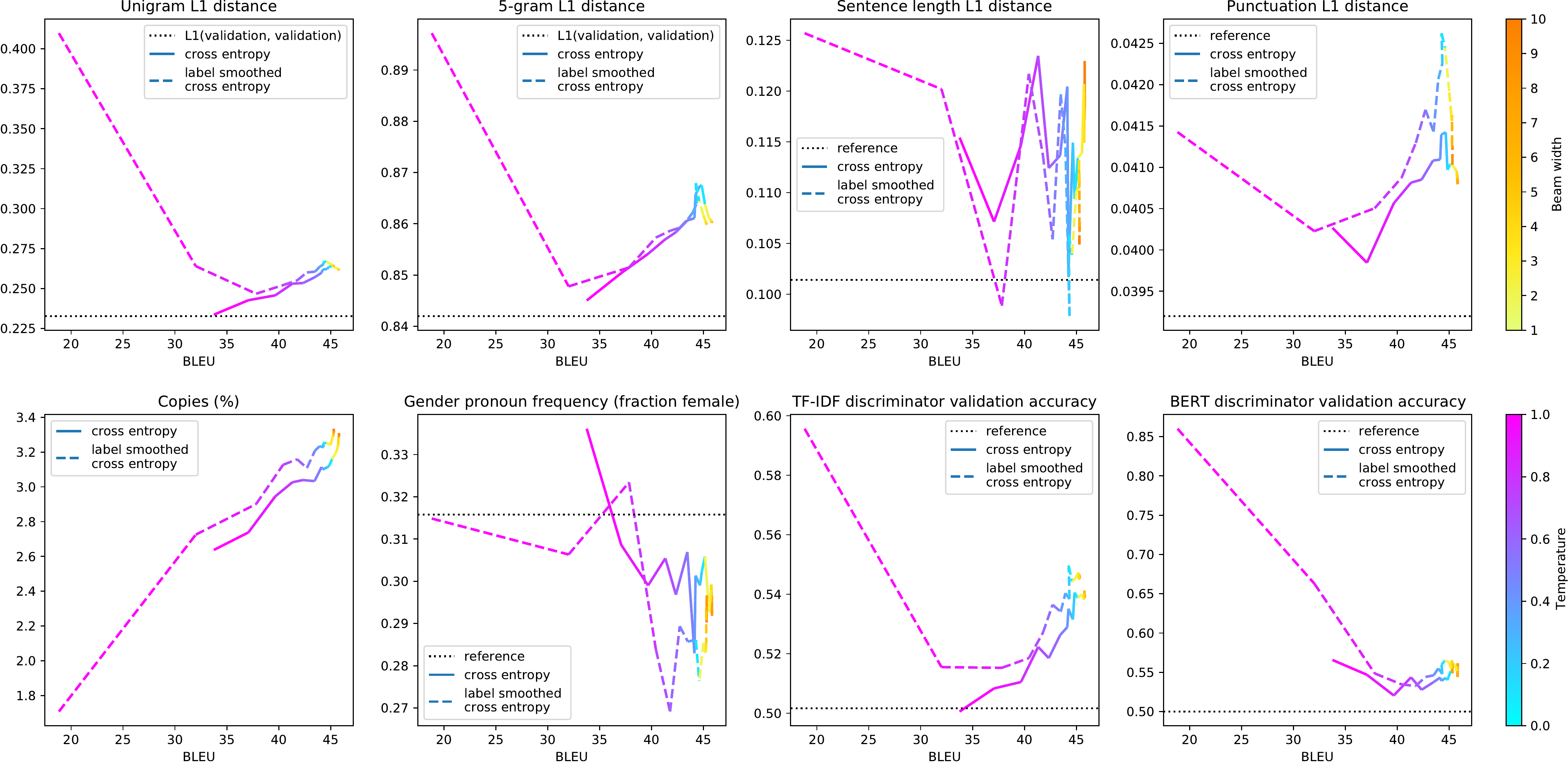}
  \caption{Results from Figure~\ref{fig:all_sweep} reproduced on the WMT 2014 English-French dataset, where the task is translation from French to English. }
  \vspace{-10px}
  \label{fig:all_sweep_fr_en}
\end{figure*}



\end{document}